%% file: root.tex
\documentclass[a4paper, conference]{IEEEtran}

\IEEEoverridecommandlockouts

\let\labelindent\relax

\usepackage{times}
\usepackage{soul}
\usepackage{url}
\usepackage[hidelinks]{hyperref}
\usepackage[utf8]{inputenc}
\usepackage[small]{caption}
\usepackage{graphicx}
\usepackage{amsmath}
\usepackage{amssymb}
\usepackage{booktabs}
\usepackage{algorithm}
\usepackage{comment}
\usepackage{multirow}
\usepackage[inline]{enumitem}

\usepackage{cite}

\usepackage{soul}

\title{\LARGE \bf Implementing BDI Continual Temporal Planning for Robotic Agents
\thanks{M. Robol and M. Roveri are partially supported by the project MUR PRIN 2020 - RIPER - Resilient AI-Based Self-Programming and Strategic Reasoning - CUP  E63C22000400001. M. Roveri and P. Giorgini are partially supported by the PNRR project FAIR - Future AI Research (PE00000013),  under the NRRP MUR
program funded by NextGenerationEU.}
}

\author{
\IEEEauthorblockN{Alex Zanetti\IEEEauthorrefmark{2}, Devis Dal Moro\IEEEauthorrefmark{1}, Redi Vreto\IEEEauthorrefmark{1}, Marco Robol\IEEEauthorrefmark{1}, Marco Roveri\IEEEauthorrefmark{1}, and Paolo Giorgini\IEEEauthorrefmark{1}}
\IEEEauthorblockA{\IEEEauthorrefmark{1}\textit{Department of Information Engineering and Computer Science - University of Trento}, Trento, Italy\\
\{alex.zanetti,redi.vreto\}@alumni.unitn.it, \{marco.robol,marco.roveri,paolo.giorgini\}@unitn.it}
\IEEEauthorblockA{\IEEEauthorrefmark{2}\textit{Eurecat - Technology Centre of Catalonia}, Catalonia, Spain \\
devis.dalmoro@alumni.unitn.it}
}

\begin{document}

\maketitle
\pagestyle{empty}

\begin{abstract}
Making autonomous agents effective in real-life applications requires the ability to decide at run-time and a high degree of adaptability to unpredictable and uncontrollable events. Reacting to events is still a fundamental ability for an agent, but it has to be boosted up with proactive behaviors that allow the agent to explore alternatives and decide at run-time for optimal solutions. This calls for a continuous planning as part of the deliberation process that makes an agent able to reconsider plans on the base of temporal constraints and changes of the environment. Online planning literature offers several approaches used to select the next action on the base of a partial exploration of the solution space. In this paper, we propose a BDI continuous temporal planning framework, where interleave planning and execution loop is used to integrate online planning with the BDI control-loop. The framework has been implemented with the ROS2 robotic framework and planning algorithms offered by JavaFF.
\end{abstract}


\section{Introduction} 
\label{sec:intro}
\input{sections/intro}

\section{Baseline} 
\label{sec:baseline}
\input{sections/baseline}

\section{A BDI architecture for continual planning}
\label{sec:cp}
\input{sections/online_planning}

\section{Implementation in ROS2-BDI} 
\label{sec:impl}
\input{sections/implementation}

\section{Validation} 
\label{sec:validation}
\input{sections/validation}

\section{Related Work} 
\label{sec:rw}
\input{sections/related_work}

\section{Conclusion and future work} 
\label{sec:conc}
\input{sections/conclusion}




\bibliographystyle{IEEETran}
\bibliography{root}

\end{document}


\maketitle

\section{Planning and search algorithms}

In AI planning~\cite{ghallab03} the search algorithms to produce a plan to achieve the goal typically search for a solution by incrementally expanding and visiting the search space driven by heuristics to direct the search towards the goal.
Such algorithms typically use an \emph{open} and a \emph{closed} list to store search nodes during the search (open list to store the nodes yet to expand, and closed to store the nodes already expanded). All these algorithms select, according to a search strategy (e.g., A*, best-first, depth-first), a node from the open list. They check whether the selected node satisfies the goal, and if it is the case, they extract and return the solution plan. Otherwise, they expand the node (e.g., by considering applicable/relevant actions depending on the search space), and add all the expanded nodes in the open list, and iterate.
This loop continues until
\begin{enumerate*}[label=(\roman*)]
\item the goal has been reached and the solution plan is given in output, or
\item no solution has been found (the open list is empty), and the non-existence of the solution is returned.
\end{enumerate*}


We propose a variant of the classical search algorithm to support the continual (temporal) planning and concurrent execution.
The proposed search algorithm is built upon a totally-ordered forward search~\cite{ghallab03}. Intuitively, we perform search iterations, where at each iteration
\begin{enumerate*}[label=\alph*)]
\item the search space is explored from the last search node from the previous iteration (the initial state in the first iteration) until a condition is satisfied (e.g., goal reached, new depth reached, number of new expanded nodes reached), and then
\item the most promising partial solution for the iteration is returned.
\end{enumerate*}

\begin{algorithm}[ht!]
\fontsize{2.9mm}{2.9mm}\selectfont
\caption{\\searchRound($mode$, $start$, $open$, $closed$)}
\begin{algorithmic}[1]
\State $best\_s \leftarrow start$
\State $open.add(start)$
\State $st \leftarrow now()$
\While{$!open.isEmpty()$}
    \State $s \leftarrow open.popMostPromisingState()$
    \If{$s.noRunActions()$ \textbf{and} $s.HVal() < best\_s.HVal()$}
        \State $best\_s \leftarrow s$
    \EndIf

    \If{$s.goalReached()$} \textbf{return} $s$
    \ElsIf{$now()-st \ge INTERVAL$} \textbf{return} $best\_s$
    \ElsIf{$best\_s.plan().size() \ge PLAN\_SIZE\_LIMIT$}
        \State \textbf{return} $best\_s$
    \EndIf
    \ForAll{$succ$ \textbf{of} $getSuccessors(getActions(s))$}
        \If{$needToVisit(succ, closed)$}
            \State $open \leftarrow searchCoreLogic(mode, succ, open)$
        \EndIf
    \EndFor
\EndWhile
\State \textbf{return} $None$
\end{algorithmic}
 \label{alg:search_round}
\end{algorithm}

The revisited version of the forward search algorithm to support continual planning supports an incremental exploration of the solution space, and the respective pseudocode is specified in Algorithm~\ref{alg:search_round}.
The algorithm takes in input a search strategy ($mode$), the current initial state of the search ($start$), and the current $open$ and $closed$ lists.
At step 1-3 the search is initialized by setting $best\_s$ (the current best state of the search) to $start$, adding $start$ to $open$, and setting $st$ to the current time instant.
Then the search starts by exploring the incrementally the search space generating new search nodes and associated plan (lines 12-14) updating $best\_s$ (line 7), and $open$ (line 14).
%
The current best state of the search is updated whenever I find a new state with improved heuristic value and not including actions started, still running and to be finished in a future state (checked through $noRunActions$ call).
The search stops under the following conditions:
\begin{enumerate*}[label=\roman*)]
\item no solution is found (line 4, open list is empty, i.e., no alternatives left to be explored);
\item the goal has been reached and the plan is given in output (line 8);
\item a given search time boundary has timed out (line 9);
\item a given maximum plan size\footnote{The plan size can be considered as the number of actions in the plan.} has been reached (lines 10-11).
\end{enumerate*}
%
This algorithm is executed as part of an outer loop (see
Algorithm~\ref{alg:search_loop}) such that at the first iteration,
both $open$ and $closed$ are empty, and $start$ is the current state
of the world from where to search for a solution plan. At each
iteration, it computes the most promising state (and related plan) or
$None$ if the search has failed to find a solution.
The $searchCoreLogic$ call at line 14 updates the open list wrt. the given search strategy ($mode$).
In this work we consider two search strategies:
\begin{enumerate*}[label=(\roman*)]
\item a greedy and fast (denoted with $G$),
\item a complete, but time consuming (denoted with $NG$).
\end{enumerate*}

\begin{algorithm}[ht!]%
\fontsize{2.9mm}{2.9mm}\selectfont
\caption{\\searchThread($improving$, $committed$)}
\begin{algorithmic}[1]

\State $open, closed, plans \leftarrow [], [], []$
\State $s\_mode \leftarrow improving$ ? $G$ : $NG$
\State $s\_prefix \leftarrow$ \textbf{await} $committed.getActions()$
\State $curr \leftarrow$  \textbf{await}  $committed.getState()$

\While{$s\_mode \neq UNSAT$ \textbf{and} $!curr.goalReached()$}
    \State $new\_state \leftarrow searchRound(s\_mode, curr, open, closed)$
    \State $commit\_p \leftarrow$ \textbf{await}  $committed.getActions()$
    \State $commit\_state \leftarrow$ \textbf{await}  $committed.getState()$
    \If{$s\_prefix.outdated(commit\_p)$} \textbf{return}

    \ElsIf{$new\_state = None$}
        \State $s\_mode \leftarrow s\_mode$ = $G$ ? $NG$ : $UNSAT$
        \If{$s\_mode = G$}
            \State $s\_prefix \leftarrow commit\_p$
            \State $curr \leftarrow init\_state$
        \EndIf
    \ElsIf{$new\_state \neq curr$}
        \If{$!improving$}
            \State $sendPartialPlan(new\_state.plan(), s\_prefix)$
        \EndIf
        \State $plans.append(new\_state.plan())$
        \State $curr, open, closed \leftarrow reroot(new\_state, open, closed)$
        \State $curr \leftarrow new\_state$
    \EndIf
\EndWhile
\If{$s\_mode$ = $UNSAT$ \textbf{and} $!improving$}
        \State $sendSearchFailed()$
\ElsIf{$improving$ \textbf{and} $improvedSolution(plans, curr)$}
         \State $sendImprovedPlan(plans, s\_prefix)$
\EndIf
\end{algorithmic}
\label{alg:search_loop}
\end{algorithm}

Algorithm~\ref{alg:search_loop} presents the search loop managing incremental exploration of the solution space performed by algorithm \ref{alg:search_round}.
The algorithm takes as input:
\begin{enumerate*}[label=\roman*)]
    \item $improving$, a boolean value set to false to quickly search for a solution, true to search for an improved one, 
    \item $committed$, a shared object that allows to retrieve a) the committed actions (via $getActions()$), and b) the committed state (via $getState()$). This structure is updated outside this search on the basis of the plan in execution.
\end{enumerate*}
The classical offline search approach is now splitted in several search iterations which incrementally move ahead in the search space, rerooting the search on the most promising state.
This algorithm runs concurrently with plan execution which, step-by-step, consumes plans feeding back plan execution state to the planner.
In the pseudo-code, we use \textbf{await} instruction to indicate the atomic reading of a shared resource updated concurrently by other threads, thus guaranteeing a consistent reading of the stored value.

At lines 1-2, $closed$ and $open$ lists and the list of $plans$ are all initialized as empty, the search $mode$ is set to Greedy ($G$) or Non-Greedy ($NG$) depending whether we are either looking for
\begin{enumerate*}[label=\roman*)]
\item a new solution from scratch to fulfill the goal or an alternative solution to avoid a possible failure in the current one, or
\item an improved solution compared to the one in execution.
\end{enumerate*}
At line 3, the set of committed actions is retrieved and stored in $s\_prefix$, which represents the prefix of the search. This will be used to understand whether the current search is outdated or not.
At line 4, the current committed state is used as initial state for the search. It represents the projection of the current state after execution of the committed actions.
At line 5, after initialization, algorithm iterates until a complete solution is found or the problem has been proved without a solution ($UNSAT$).
At line 6, the exploration of the search space is moved forward by calling Algorithm~\ref{alg:search_round}.
Note that, open and closed are updated in $searchRound$ and, if not updated before that, will be available "as-is" in the next iteration, therefore search progress is not lost.
At line 9, search is interrupted if outdated, i.e. concurrently, further actions have been committed that make the search inapplicable. The search starts from a state that came after execution of actions in the search prefix; if additional actions are committed, the search becomes outdated.
At lines 10-14, if search iteration fails to move forward, no state is returned ($None$), the open list was entirely consumed and the search is considered as failed. However, while the failure of the greedy search causes the fallback to the complete one; the failure of the latter approach causes the search to consider the problem unsatisfiable.
Lines 15-20 handle the case in which the search iteration  has moved forward.
At line 16-17, we're not looking for an improved solution, therefore, the new plan is sent along with the computed search prefix to the Intention Scheduler.
At line 18, partial plans are incrementally stored.
At line 19, search is re-rooted for next iteration, clearing out the whole closed list and filtering the open to keep only states with same plan prefix as the one in the new found plan.
Then, this plan prefix is truncated from everywhere so that next search will see the current state, effectively as the root one. 
Finally, at lines 23-24, when searching for an improved solution and the solution is verified to improve current one, it is sent to the Intention Scheduler.
If the search is not looking for improved solutions, the failure of the search is communicated (see line 22), so a connected Intention Scheduler can act accordingly.

\bibliographystyle{named}
\bibliography{references}

%% file: sections/intro.tex
Autonomous systems  are widely adopted in
 industrial environments where they are used in strictly controlled areas, such as assembly lines in the industry or automated
warehouses in logistics.
%
%
%
Along this, there is a growing interest in adopting autonomous systems in more
complex and highly dynamic applications in which they
are required to deal with unexpected events (e.g, the interaction with
humans or with other robotic systems), or changes in objectives (e.g.,
deal with new high-priority activities).
This demands fast adaptation to promptly respond to the
perceived changes, and to take decisions and actuate them having not
yet performed complete reasoning on the problem, but having only
explored partially the solution.

Recent frameworks enabling a robotic agent to autonomously deliberate have been proposed~\cite{DBLP:journals/aepia/GottifrediTCGS10,breemen2013,duffy1999social,DBLP:conf/woa/AlzettaG19,cognitao,DBLP:journals/cogsr/BustosMBRGM19,DBLP:conf/taros/PolydorosGRNK16,DBLP:conf/aips/CashmoreFLMRCPH15,DBLP:conf/iros/0001CMR21}, but they
are still in an early stage and present a number of open problems
(e.g., they provide planning functionalities, but they are not
supported by any deliberative functionalities making planning unusable
in real scenarios).
\cite{PAAMS22} proposed a BDI
(Belief-Desire-Intention)~\cite{DBLP:conf/kr/RaoG91} architecture,
called ROS2-BDI, able to use the beliefs of an agent to reason about
goals and elaborate plans to achieve them. The proposed
architecture supports
\begin{enumerate*}[label=\roman*)]
\item BDI-based deliberation to develop agents with temporal planning
  capabilities;
\item deadline-aware prioritization of desires;
\item preemption of running plans with lower priority.
\end{enumerate*}
This architecture, despite being able to generate new plans in
response to dynamic changes in the environment suffers from the problem
that before starting any actions, the agents shall generate a full plan
to achieve its goals. Moreover, it may exhibit continuous run-time
failure and subsequent replanning
to try to adapt to the contingencies thus preventing fruitful progress
towards the goal.
In the setting of AI planning it has been studied the problem of
generating more robust plans able to adapt to a set of
contingencies~\cite{rt_concurr_pexec_stoc_domains,DBLP:journals/ai/BertoliCRT06}, the problem of continual planning
while acting~\cite{concurr_pexec_arobot,895924,9473021} or interleaving planning and
execution~\cite{DBLP:journals/ai/PatraMGNT21,cont_planning_acting_ma}.
However, all these approaches are rather limited: they assume a
classical plan-act
sequence~\cite{rt_concurr_pexec_stoc_domains,DBLP:journals/ai/BertoliCRT06},
consider only motion-planning while acting
\cite{concurr_pexec_arobot,895924,9473021}, need to specify when and
how to switch from planning to
execution~\cite{DBLP:journals/ai/PatraMGNT21,cont_planning_acting_ma},
do not show temporal planning capabilities (a fundamental element for
continual planning in realistic scenarios).

In this paper, we make the following contributions.
First, we specify a new BDI architecture that extends
ROS2-BDI~\cite{PAAMS22} to
\begin{enumerate*}[label=\roman*)]
\item allow an agent to take decisions and actuate them having not yet
  computed a complete plan for the active goals, but only
  explored partially the solution;
\item handle preemption of running plans to deal with changes in the
  objectives and/or to execute the new plan or the remaining partial
  plan towards the goal computed while executing the previous one.
\end{enumerate*}
Second, we specify how to modify a state-of-the-art temporal planning
algorithm to support continual temporal planning (provide intermediate
candidate solutions to execute, while continuing the search).
Third, we implemented the continual temporal planning algorithm in a
new temporal planner, named CJFF, built on top of
JavaFF~\cite{javaff}. The novel framework has been implemented on top
of the ROS2-BDI leveraging our CJFF.
Finally, we demonstrate the practical applicability of the novel implemented framework through its application in a highly dynamic
resource collection robotic scenario where we compare the approach of
ROS2-BDI baseline with our extended ROS2-BDI framework. The results show
that on average the new approach results in shorter plans w.r.t. the
baseline, thus indicating the ability of the new framework to adapt
more easily to contingencies.


The paper is organized as follows. In Sec.~\ref{sec:baseline} we
briefly present the baseline. In Sec.~\ref{sec:cp} we describe the
novel BDI architecture and the algorithm for continual temporal
planning. In Sec.~\ref{sec:impl} we discuss the implementation within
the ROS2-BDI framework, and in Sec.~\ref{sec:validation} we present the
results of the validation of the proposed framework. In
Sec.~\ref{sec:rw} we discuss the related work, and finally, in
Sec.~\ref{sec:conc} we draw conclusions and outline future works.

%% file: sections/baseline.tex
In this work, we build on
\begin{enumerate*}[label=\roman*)]
\item JavaFF, a PDDL 2.1 temporal planner developed in Java;
\item ROS2-BDI, a BDI framework implemented on top of ROS2.
\end{enumerate*}
In the rest of this section, we summarize the respective main concepts.

\begin{figure}[t!]
  \centering
  \vspace{5pt}
  \includegraphics[width=1\linewidth]{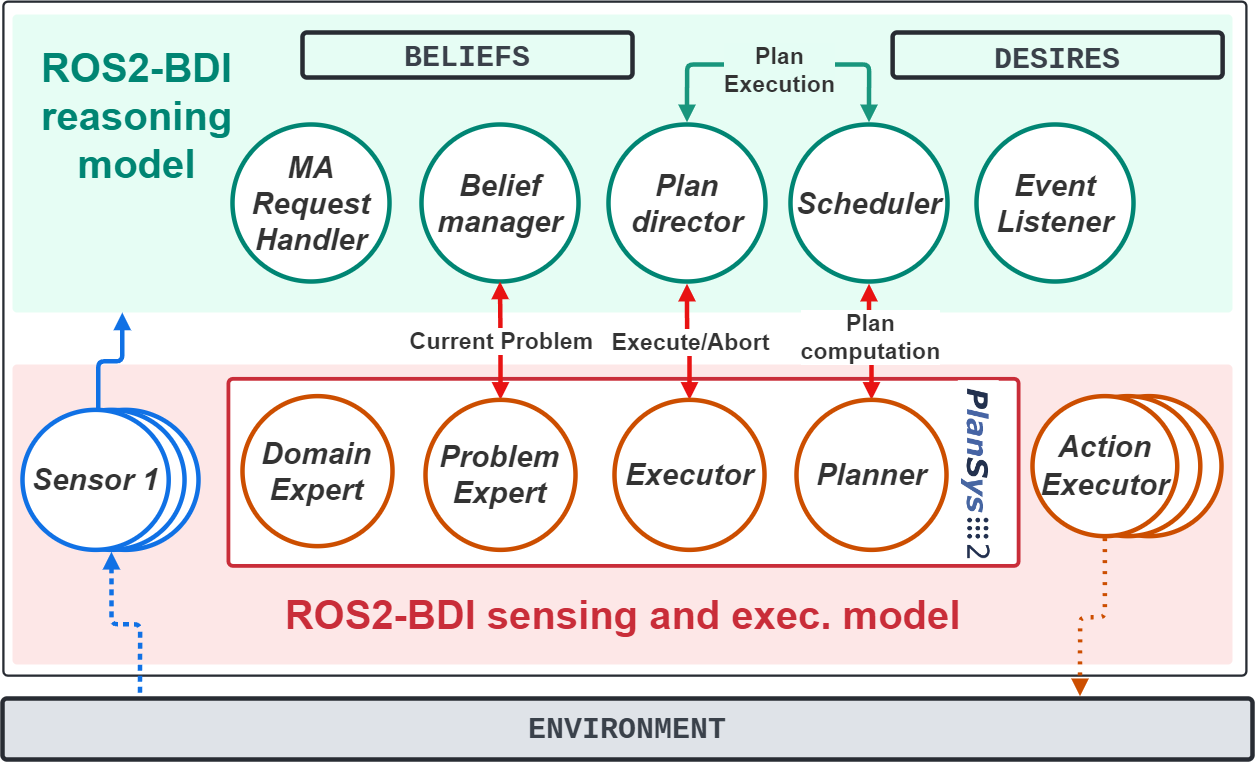}
  \caption{Architecture of a ROS2-BDI agent. Circles are ROS2 nodes,
    arrows and buses (rectangles) represent communications.}
  \label{fig:ROS2-BDI}
\end{figure}

\textbf{PDDL 2.1}~\cite{DBLP:journals/jair/FoxL03} is the "standard"
formalism adopted for modelling the knowledge and the behaviour of
agents considering that actions might last some known amount of
time. In PDDL2.1, the effects and conditions for each action might apply
before, during and/or after its execution. PDDL 2.1 compliant planners
are capable of computing time-triggered plans to go from an initial
state to a goal one, each action is associated with a start time and a
given duration, and multiple actions can occur concurrently.
\textbf{JavaFF}~\cite{javaff} is a re-implementation for didactic
purposes of the MetricFF~\cite{DBLP:journals/corr/abs-1106-5271} in
Java, with also support for PDDL2.1 temporal planning. Being developed
for didactic purposes, its heuristic and performance did not age well
compared to other state-of-the-art PDDL 2.1-compliant planners, such
as POPF~\cite{DBLP:conf/aips/ColesCFL10} or
OPTIC~\cite{DBLP:conf/aips/BentonCC12}. However, the underlying framework is robust, modular, and relatively easy to modify and build
on w.r.t. other more efficient temporal planners.

\textbf{ROS2-BDI}~\cite{PAAMS22} is a BDI
framework for developing distributed autonomous robotic systems built
on top of PlanSys2~\cite{DBLP:conf/iros/0001CMR21} within
ROS2~\cite{ROS2}. It supports BDI-based deliberation through the
generation of time-triggered temporal plans thanks to the
POPF~\cite{DBLP:conf/aips/ColesCFL10} integrated within PlanSys2.
Figure \ref{fig:ROS2-BDI} depicts the architecture of a ROS2-BDI
agent. ROS2 \emph{nodes} (depicted as circles) encapsulate the core
functionalities of the agent. The communication of nodes within an
agent and among agents happens through communication channels (named
\emph{topics} within ROS2).
ROS2-BDI core nodes mainly deliver the following features:
\begin{enumerate*}[label=\roman*)]
\item Belief Management
\item Multi-Agent Requests handler
\item Scheduler (desire prioritization and preemption)
\item Check over the running plan's context and desired deadline conditions
\item Event Listener
\end{enumerate*}.
The Scheduler node provides the handling of prioritization of running
intentions and goals, interfacing with a) the
PlanSys2~\cite{DBLP:conf/iros/0001CMR21} reference framework for PDDL
2.1 temporal planning within ROS2, b) and with the Plan Director to
demand either their execution or abortion.  Given a plan along with
its preconditions, desired deadline and context conditions, the plan
director is used to trigger, monitor and abort its execution. Once a
desire is activated and a plan for fulfilling it is demanded for
execution, it is run through its termination: either by failure or
success.  Plans and action executions strongly rely on
PlanSys2. Finally, the Event Listener enforces the belief revision and
option generation functions for the ROS2-BDI agent.

%% file: sections/online_planning.tex
%
%

%
The BDI model is progressively adopted into autonomous agents to mimic
human behaviour and make them able to deal with complex problems by adapting their reasoning and acting to changes happening in their environment while operating.
A set of plans is provided in their knowledge base to reach the desires
with alternative solutions.
Planning capabilities allow agents to synthesize new plans to reach a
given goal starting from given initial conditions. Only when a complete
plan has been computed, the agent starts its execution.
However, when the environment changes repeatedly, the capability to
plan a sequence of actions to perform while acting (known as
\emph{continual planning}) is of primary importance to leverage
opportunities and tackle contingencies of the evolving scenarios.
Agents featuring continual planning and execution can
\begin{enumerate*}[label=\roman*)]
\item start executing a plan when it is not yet complete, focusing
  first on the immediate things perform, and
\item adapt their course of actions along with the changes in the
  environment and in the goals they're trying to pursue.
\end{enumerate*}
In the following, first, we describe how to extend a BDI architecture
equipped with deliberation capabilities to support continual
temporal planning, and then we illustrate the most relevant features needed for smooth
integration. Namely,
\begin{enumerate*}[label=\roman*)]
\item plan failure forecasting,
\item run-time goal revision,
\item rescheduling in case of an improved solution is found.
\end{enumerate*}

\subsection{Reasoning and acting framework}
\label{ssec:pexec_loop}

\begin{figure}[t!]
  \centering
  \vspace{5pt}
  \includegraphics[width=1\linewidth]{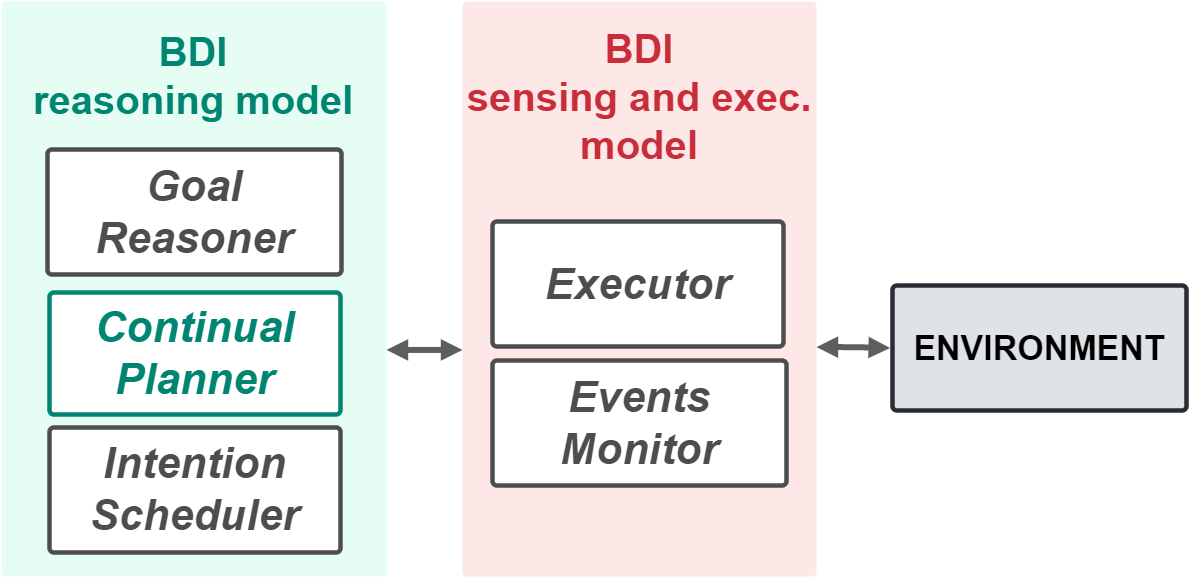}
  \caption{Architecture of the reasoning and acting framework.}
  \label{fig:loop}
\end{figure}

Fig.~\ref{fig:loop} shows the core architectural components of the proposed BDI model with continual planning and concurrent execution.
On the left, the BDI reasoning model includes the \emph{Continual Planner}, the \emph{Goal Reasoner}, and the \emph{Intention Scheduler}.
In the middle, the BDI sensing and execution model includes the \emph{Executor} and the \emph{Events Monitor}, respectively acting and sensing to/from the \emph{Environment}.


The \textbf{Goal Reasoner}, given the agent behavioural model, creates new goals and pushes them in the desire set.
Then, it activates a goal (from the desire set), selected on the basis of preconditions and priorities.
The \textbf{Continual Planner} redefines the considered planning problem for each received goal, taking into account updates about the state of the agent and of the world. Solutions are continuously produced accordingly.
The Continual Planner adopts a novel search approach, which allows the exploration of the solution space step-by-step.
Intuitively, the search is performed considering a limited search horizon, so that, at each iteration, it incrementally refines the previous plan adding new actions towards the goal and/or completely revising the plan to handle contingencies.
Therefore, partial plans produced by each search iteration do not necessarily bring to the final goal.
Classical temporal planning heuristics are used by the Continual Planner while computing each partial plan to choose among the most promising alternative actions (see Sect.~\ref{ssec:planning_search} for more details).
%
The \textbf{Intention Scheduler} is responsible for mantaining a queue of plans as received by the Continual Planner, for their execution by the Executor.
%
A plan can either be
\begin{enumerate*}[label=\roman*)]
\item a continuation of a previous plan, or
\item an alternative (e.g., a better solution) which might imply deviating from current execution and discarding previously queued plans.
\end{enumerate*}
Respectively, those lead to
\begin{enumerate*}[label=\roman*)]
\item enqueueing of the received incremental plan or
\item replacement of all the plans in the queue with the new one. In this last case, it might also tell the Executor to arrest earlier the plan in execution at the point at which the new plan will apply.
\end{enumerate*}
%
%
The \textbf{Executor} is responsible for
\begin{enumerate*}[label=\roman*)]
\item the execution of a plan by checking the applicability of each of the actions within the plan before dispatching the execution of the
  action itself;
\item monitoring the correct termination of the dispatched actions to decide whether to proceed with the execution of the remaining actions in the plan or stop it and notify the failure to activate further reasoning;
\item interrupting the execution of the current plan when requested by the Intention Scheduler.
\end{enumerate*}
%
%
The \textbf{Events Monitor} senses the environment and actions
execution status and updates the knowledge of the agent accordingly.

\begin{figure}[t!]
  \centering
  \vspace{5pt}
  \includegraphics[width=1\linewidth]{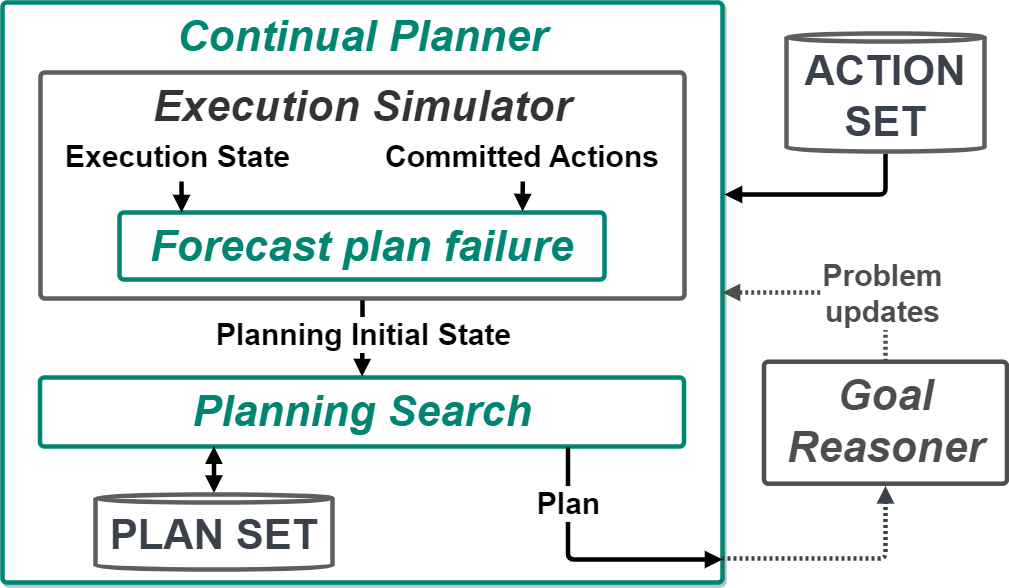}
  \caption{Details of the \textbf{Continual Planner}.}
  \label{fig:planner}
\end{figure}

Fig.~\ref{fig:planner} provides details of the \textbf{Continual Planner}, here represented as composed by two sub-components: an \emph{Execution Simulator} and a \emph{Planning Search}.
The \textbf{Execution Simulator} receives updates from the Events Monitor and consequently updates its internal representation about the \textit{Execution State} of plans (running, successful, failed) including \textit{Committed Actions}, part of the plan that it is committed to executing, also considering additional constraints (e.g., the minimum number of committed steps).
The \textbf{Forecast Plan Failure} component of the Execution Simulator provides forecasting capabilities to verify whether the current plan is going to fail in its execution. This is performed by progressing the current state of the world guided by the plan (considering their pre-conditions and effects). If for some reason (e.g., a precondition of an action in the plan does not hold), then it informs the Planning Search to find another alternative plan. If the simulation of the plan succeeds, then the progressed state (named \textit{Planning Initial State}) is sent to the Planning Search for continuing the previous search.
The \textbf{Planning Search} component receives in input the \textit{Planning Initial State}, the goal to achieve, and the set of actions to orchestrate to build a solution plan.
%
In Sec.~\ref{ssec:planning_search}, we  detail how to modify a temporal planning search to behave as discussed above.



\subsection{Search components for continual planning}
\label{ssec:planning_search}



In AI planning, the search for a plan to achieve the goal is performed by incrementally by expanding and visiting the search space (see ~\cite{ghallab03} for details). Heuristics can be adopted to drive the search to converge faster towards the goal.
Such searches typically use algorithms based on an \emph{open} and a \emph{closed} list to store nodes of the search space during the search (open list to store the states yet to expand, and closed to store the states already expanded). All these search algorithms, according to a a given expansion strategy (e.g., A*, best-first, depth-first),
\begin{enumerate*}[label=(\roman*)]
    \item select a state from the open list,
    \item expand it (e.g., by considering applicable/relevant actions),
    \item add all the generated states in the open list and
    \item iterate until the goal is reached or no solution is found (open list is empty);
\end{enumerate*}
Finally, a solution plan is given in output or the non-existence of the solution is returned.

We propose a variant of the classical search to support execution at planning time, in a concurrent, temporal, round-based and continual fashion.
%
The proposed search is built upon a classical offline totally-ordered forward search~\cite{ghallab03}, now splitted into \textbf{search rounds} which incrementally move ahead in the search space. As soon as the first search round produces a partial plan, execution can start without having to wait for the complete plan.
Intuitively, each round executes only a limited number of expansion iterations before exiting to then resuming from the most promising state:
\begin{enumerate*}[label=\alph*)]
    \item search starts/resumes from the last round most promising state, then
    \item expansion strategy is applied and open and closed lists are updated appropriately and
    \item iterations continue until a round-termination condition is satisfied (e.g., goal reached, new depth reached, number of newly expanded nodes reached)
\end{enumerate*}
then the most promising state is returned by the search round.
\textbf{Re-rooting} happens between rounds, clearing out the whole closed list and filtering the open to keep only states with the same plan prefix as the one in the new found plan. Then, this plan prefix is truncated from everywhere so that the next search will see the current state, effectively as the root one.
To compute the plan as soon as possible, a greedy expansion strategy is initially adopted. Still, in the case of no solution found, the search switches to a non-greedy search used as a \textbf{fallback expansion strategy}.
%
The search runs concurrently with plan execution which, step-by-step, consumes plans and \textbf{commit for execution}.
When a plan is committed for execution, the expectation is to reach the projection of the current state after the plan, since the request for abortion of commitment is not allowed.
In the case of failure in the execution of actions, a goal change, or an improved plan, \textbf{continual search revision} re-configure the search to adapt. The search is reinitialized from the committed projection state. When the plan is produced, the old plan, queued but yet not committed for execution, is early aborted and replaced with the new plan. However, if the search does not produce the new plan within the time window covering the current execution commitment, the search becomes outdated and needs to be reinitialized again from the newly committed projection state.

\paragraph{Continual goal revision.}
\label{ssec:rt_goal_augment}

In a BDI agent, goals are fulfilled separately, for example, with plans executed in sequence. However, optimization may be possible, e.g. an overall-shorter plan may exists that fulfill multiple goals. Still, goals are pushed to the agent at execution time, for example after observation of the environment, when the fulfillment of a previous goal has already started. Under this condition, a revision of currently active goal is needed to generate an overall-shorter new plan across goals. 
The Goal Reasoner component (see Fig.~\ref{fig:loop}) supports continual goal revision, based on domain-specific user-specified rules, so that current goal could be revised to include new ones, then fulfilled with a possible overall-shorter plan.
So far we support a very simple rule that computes a revised goal as $G'' = G \cap G'$ that considers the old goal $G$ and new goal $G'$.
Since $G''$ is more specific than its base version $G$, the revision of a goal may not be always affective. A rules-based mechanism is used to specify the conditions for revision, goal-by-goal.

\paragraph{Improve solution.}
\label{ssec:improve_solution}

However, a complete plan may finally be found earlier before the execution terminates, so that there is still time for the planner to improve the solution.
Our framework adopts two efficiently-different searches, first, a quick greedy search looks for the first viable solution, and then, a non-greedy search tries to optimize it.
If an improved plan is found, it is scheduled in the waiting queue, by requesting an early arrest of all scheduled actions diverging from the improved plan.
To be schedulable, when produced, the plan must be non-outdated with respect to the updated execution state.
%


\subsection{Forecasting plan failure}
\label{ssec:forecast_plan_failure_goal_sim}


Uncontrollable events that may happen in the environment can cause failures in the execution of plans.
In classical planning agents, a plan execution failure directly triggers a re-planning.
Our architecture continuously simulates the execution of the current plan from the current state, so to forecast a possible failure.
The simulation consists in progressing the current state by applying the actions in the plan, checking if the action preconditions hold, and then updating the state with the action's effects, otherwise reporting failure. Failure is also reported if the plan does not allow goal achievement.
Whenever a plan simulation fails, two cases need to be considered.
The failure is forecasted to occur within the committed actions: in this case, nothing can be done to avoid the natural failure of the plan.
The failure is forecasted to occur after the committed actions: in this case, an alternative plan is generated by initializing a new search starting from a committed state (the state resulting from the simulation of the committed actions from the current state).
%

%% file: sections/implementation.tex
We implemented the proposed framework on top of ROS2-BDI~\cite{PAAMS22}.
To this extent, we first extended Java-FF to support the proposed
continual temporal planning, and this results in a new tool named CJFF~\footnote{CJFF can be found at \url{https://github.com/RTI-BDI/JavaFF}.};
then we extended ROS2-BDI~\cite{PAAMS22} to support the
proposed continual planning reasoning~\footnote{The extended ROS2-BDI framework is available at \url{https://github.com/RTI-BDI/ROS2-BDI-ONLINE/tree/ojff}.};
in addition, we also revised PlanSys2, to support continual planning execution, and adopted it as ROS2 planning system~\footnote{The adopted ROS2 planning system, a revised version of PlanSys2, is available at \url{https://github.com/RTI-BDI/ros2_planning_system}.}.
%

\paragraph{Continual JavaFF.}
\label{ssec:ojff}

We implemented the proposed continual planning algorithm sketched in
the previous sections in a new planner called CJFF (Continual
JavaFF). CJFF builds on and extends JavaFF~\cite{javaff} to fulfil
the need for a general-purpose, PDDL 2.1 compliant, temporal planner.
JavaFF, although it offers obsolete heuristic w.r.t. other state-of-the-art temporal planners, offers a totally-ordered forward chaining
approach, its implementation is clear, modular and flexible, thus
making it a good fit for the first implementation of our planning search.
For CJFF we leverage the two search strategies provided by JavaFF,
namely greedy Enforced Hill Climbing (EHC) and Best First Search (BFS)
both based on a Relaxed Plan Graph (RPG) heuristic extended to deal
with temporal actions~\cite{javaff}.

We wrapped CJFF into a RCLJava~\cite{rcljava} ROS2 node, so to support
its integration into a ROS2 framework.
The node consists of two main threads:
\begin{enumerate*}[label=\roman*)]
\item a control thread that communicates with the other ROS2 nodes of
  the architecture through specific ROS2 topics to enable monitoring
  search progresses and controls planning and replanning phases;
\item a search thread responsible for the search.
\end{enumerate*}
CJFF exposes ROS2 services to trigger a new search from scratch (using
an initial state computed directly via the PDDL-encoded problem).
CJFF also publishes its search progresses i.e., ordered incremental
plans, where the plan baseline specifies the situation in which each
incremental plan can be executed.
In addition, CJFF expects to get execution status information to
update the plan execution state and committed status.
CJFF computes committable states of the plan where no open action
exists\footnote{The search performed by JavaFF to compute a time-triggered plan leverages a reduction to classical planning by
  creating a classical planning problem where each temporal action is
  encoded with so-called
  snap-actions~\cite{DBLP:journals/ai/ColesFHLS09}. Once a classical
  planning solution is computed, we need to check the consistency of
  the temporal network induced by the computed classical plan. If
  consistent, then a time-triggered plan is extracted. Otherwise, the
  search continues. We refer the reader
  to~\cite{DBLP:journals/ai/ColesFHLS09} for further details.}%
, considering also a minimum number of actions specified by the user.
Computation of the "projected state" that the agent is committed to reaching is performed via simulation, applying the expected effects of
the upcoming actions in the executing plan. The same mechanism is used to
forecast plan failure, as detailed in
Sect.~\ref{ssec:forecast_plan_failure_goal_sim}.

\paragraph{Extending ROS2-BDI architecture.}
We implemented the high-level architecture described in
Sect.~\ref{sec:cp} to support continual planning building on top and
extending the ROS2-BDI~\cite{PAAMS22} with the CJFF functionalities.
\begin{figure}[t!]
  \centering
  \vspace{5pt}
  \includegraphics[width=0.9\linewidth]{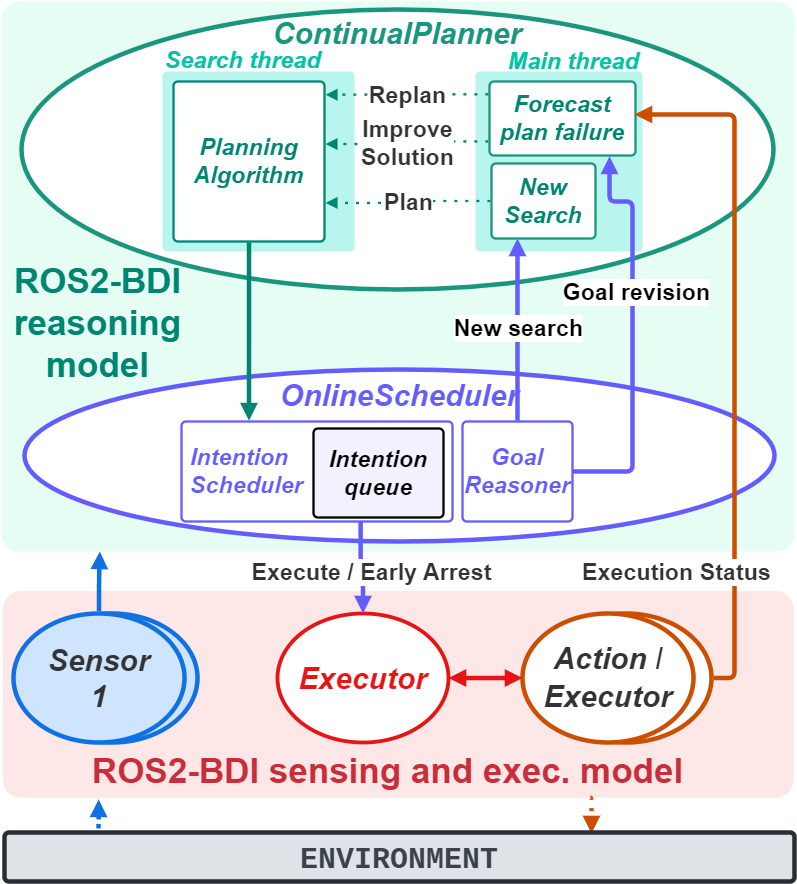}
  \caption{The ROS2-BDI architecture for continual planning.}
  \label{fig:implementation}
\end{figure}
The resulting ROS2 architecture is depicted in
Fig.~\ref{fig:implementation}.
The architectural components of Fig. \ref{fig:loop} and
\ref{fig:planner} are here implemented into ROS2 nodes, modifying and
extending ROS2-BDI core nodes, as follows:
\begin{enumerate*}[label=\roman*)]
\item \textbf{CJFF} replaces the PlanSys2 ROS2-BDI component for
  offline planning;
\item \textbf{Online Scheduler} replaces the ROS2-BDI Scheduler;
\item \textbf{Executor}, still based on PlanSys2, has been
  re-implemented to allow for the safe termination of the execution of a
  plan (early arrest).
\end{enumerate*}
The original ROS2-BDI scheduler activates a goal only when provided
with a complete solution plan for the goal itself, and after
verifying deadlines.
CJFF may not provide a complete plan all at once, thus the Online Scheduler shall activate the goal without knowing whether a complete
plan exists.
Similarly to ROS2-BDI Scheduler, it checks whether the precondition
holds, and deals with goal priorities.
The Online Scheduler initializes CJFF to start a new search as soon as
a goal has been activated. Then it waits for the search  progresses (e.g., give possibly new partial plans).
CJFF generates and publishes the new possibly partial plans as a
continuation of the previously computed ones, or a new plan to be used
in place of the one currently in execution.
The Intention Scheduler monitors the CJFF node to decide how to
proceed, i.e., enqueue the new partial plan, or ask the Executor to
safely terminate the plan in execution, and then proceed with the new
one (see Sect.~\ref{sec:cp}).
The Online Scheduler is also responsible for revising the goals within
the Gaol Reasoner as explained in Sect.~\ref{ssec:rt_goal_augment}.
The preemptive mechanism of the original ROS2-BDI Scheduler was also
modified to request the Executor to safely terminate the plan in
execution before switching to the activation of a higher priority
desire.
The Executor, based on PlanSys2, has been also thoroughly modified to
support the safe termination of the plan currently in execution. In
particular, let $\pi$ be a time-triggered plan currently in execution
containing an action \texttt{A}. The request of the safe termination
after \texttt{A} requires waiting also the termination of all the
other actions \texttt{B\textsubscript{i}} $\in \pi$ whose execution
started before \texttt{A} naturally finishes.
Finally, each Action Executor publishes the respective execution
status (e.g., running, success, fail) so that other ROS2 modules, e.g.,
the CJFF node can intercept to decide what to do next.

%% file: sections/validation.tex

In order to validate our novel approach, we conceived a scenario where
there are several recycling agents able to move in a 2d grid map with static
and known a priori obstacles (e.g., walls), and (pseudo) randomly
moving obstacles (e.g., people) whose movements are not known a priori, and whose presence can only be discovered through sensing
while moving (the agent can sense free locations within a given
distance from its position).
Each agent can \texttt{move} (from one cell to an adjacent one, known
to be \texttt{free}), \texttt{pickup} a litter, and \texttt{recycle}
it. All these actions are durative actions with an associated known
duration.
Each agent has the desire to collect and dispose of all the litter he
discovers in the nearest bin, and shall not move to a location
occupied by one of the moving obstacles or by another agent. New
litters may appear randomly in the grid and contribute to changing the goal of the agents when discovered.

\begin{figure}[tb]
  \centering
  \vspace{5pt}
  \includegraphics[width=1\linewidth]{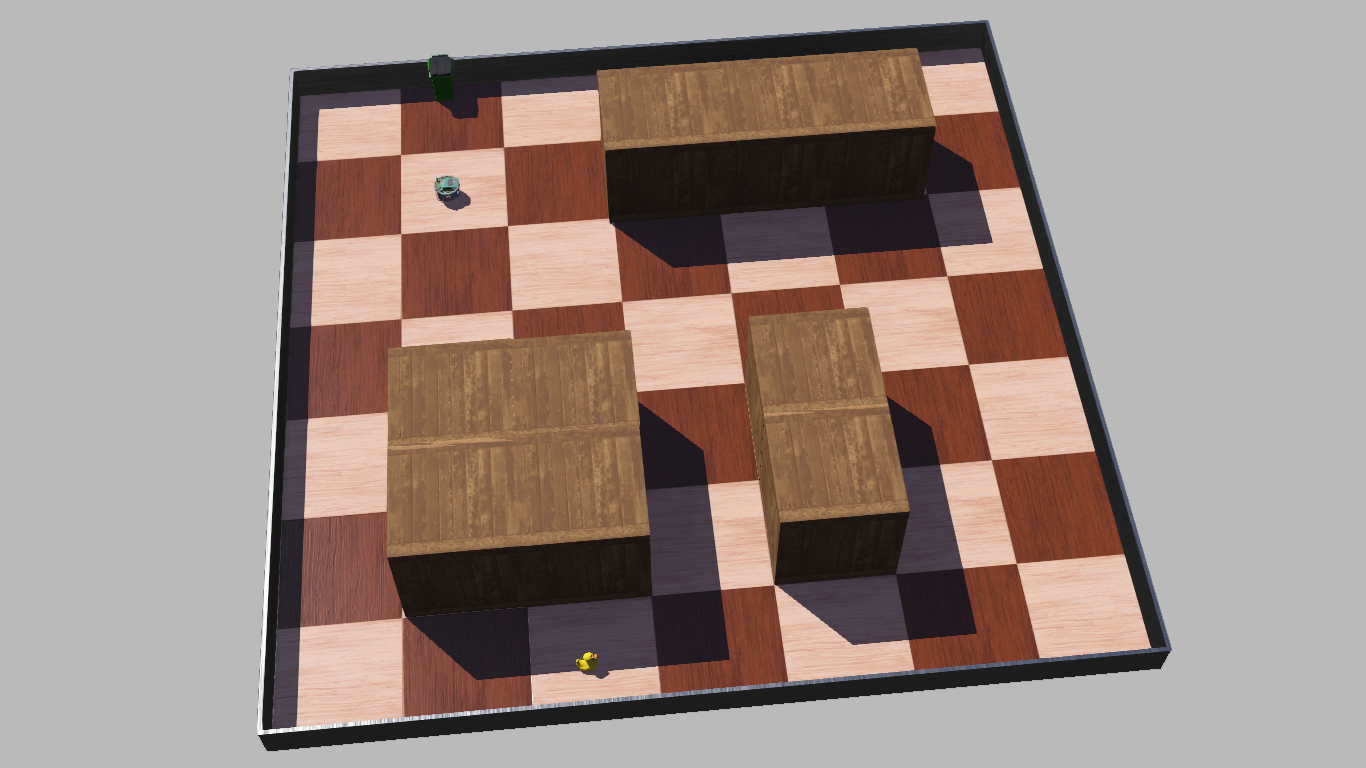}
  \caption{Validation scenario implemented in the Webots simulator.}
  \label{fig:demo}
\end{figure}

Fig.~\ref{fig:demo} shows the validation scenario, in which the recycling agent (here represented by the robot) move on the grid to collect garbage (here represented by ducks).
The validation scenario has been implemented in a ROS2-based simulation environment~\footnote{The implementation of the validation scenario based on Webots is available at \url{https://github.com/RTI-BDI/Redi-Webots-scenarios}.}.
Specifically, simulation environment is provided by Webots, a simulator for robotic systems that comes with ready-to-use robot models, already bundled with the ROS2 protocol~\footnote{More on Webots can be found at \url{https://cyberbotics.com/}.}. Being implemented on top of ROS2, these same experiments could be easily repeated with real robots with no additional coding or configuration, except for minor settings.
We also created a corresponding PDDL 2.1 domain specifying the
predicates needed to encode the scenario (e.g., \texttt{free(x,y)}
and \texttt{litter(x,y)} to indicate resp. whether location
\texttt{x,y} is free and the presence of a litter in \texttt{x,y}),
the different actions (e.g., \texttt{move}), and \texttt{pickup}).

For the validation, we considered a simple 7x7 map with the static obstacles positioned as per Fig.~\ref{fig:demo}~\footnote{A video of the validation scenario can be found at \url{https://github.com/RTI-BDI/iros2023/raw/main/on_moving_SR_400_LS_800_DC_4.mkv}.}.
Because of a low-level bug in PlanSys2 related with the construction of Behavior Trees, we restricted the analysis to one single agent.
To compare the new ROS2-BDI framework with the old one we created
pseudo-random simulations where there is one recycling agent, two
persons that will follow a predefined circular path (not known by the
agent) each performing a step forward along the respective path for
$\mathtt{move} \in \{4,8\}$ times, three litter items placed in the
map, however only the position of $\mathtt{k\_litt} \in \{2,3\}$ are
known by the agent. We also considered different sizes
$\mathtt{det\_area} \in \{1,2,3\}$ of the detection area of the
agent. (An assignment to \texttt{move,det\_area,k\_litt} defines a
\emph{situation}.)
We measure the number of steps required by the recycling agent to
dispose of all the placed litter items on the map.
For a fair comparison of the new ROS2-BDI framework and the previous
one, we integrated into the old ROS2-BDI framework JavaFF, such that
it can be used in offline runs, replacing the default (more
efficient) POPF planner used within PlanSys2.
Finally, for each situation, we performed 4 runs and collected the average number of moves of the recycling agent to dispose of all the
litter items.
We remark that, a comparison of computation time between the original
ROS2-BDI framework and the one proposed in this paper, would not be
fair because here we care about plan quality (number of steps
performed by the recycling agent) rather than on the time to compute
plans (that depends on the implementation language and heuristic used
for the search).
We run all the tests on a notebook equipped with an
Intel\textcopyright{} 2.80GHz i7\texttrademark{} CPU with 16GB of RAM,
running Linux.

The results are reported in Table~\ref{tab:eval}. The first column
reports the considered parameter setting
(\texttt{move,det\_area,k\_litt}), the second and fourth columns are
the average ($\mu$) number of steps performed by the recycling agent
in four runs, the third and fifth columns are the respective standard
deviation. Finally, the sixth columns ($\mu+$) is the difference of
the second and fourth columns (it represents the average number of
additional steps performed by the offline approach).
\begin{table}
  \fontsize{2,5mm}{3mm}\selectfont
  \centering
  \vspace{5pt}
  \begin{tabular}{|c|c|c|c|c|c|}
    \hline
    \multirow{2}{*}{Setup} & \multicolumn{2}{|c|}{Offline}& \multicolumn{2}{|c|}{Online} & \multirow{2}{*}{$\mu+$} \\
    \cline{2-5}                    &  $\mu$ & $\sigma$ & $\mu$  & $\sigma$&        \\ \hline
    move=4, det\_area=1, k\_litt=2 & $56.0$ & $2.3$    & $42.5$ & $9.8$   & $13.5$ \\ \hline
    move=8, det\_area=1, k\_litt=2 & $58.5$ & $6.8$    & $43.0$ & $10.1$  & $15.5$ \\ \hline
    move=4, det\_area=2, k\_litt=2 & $58.5$ & $3.3$    & $36.0$ & $0.0$   & $22.5$ \\ \hline
    move=8, det\_area=2, k\_litt=2 & $58.0$ & $5.9$    & $35.5$ & $6.0$   & $22.5$ \\ \hline
    move=4, det\_area=3, k\_litt=2 & $66.0$ & $4.0$    & $42.5$ & $10.5$  & $23.5$ \\ \hline
    move=8, det\_area=3, k\_litt=2 & $57.0$ & $4.2$    & $42.0$ & $6.3$   & $15.0$ \\ \hline \hline
    move=4, det\_area=1, k\_litt=3 & $48.0$ & $0.0$    & $43.5$ & $5.5$   & $4.5$  \\ \hline
    move=8, det\_area=1, k\_litt=3 & $57.8$ & $0.5$    & $36.0$ & $2.8$   & $21.8$ \\ \hline
    move=4, det\_area=2, k\_litt=3 & $48.0$ & $0.0$    & $35.5$ & $2.5$   & $12.5$ \\ \hline
    move=8, det\_area=2, k\_litt=3 & $57.0$ & $1.2$    & $37.0$ & $11.6$  & $20.0$ \\ \hline
    move=4, det\_area=3, k\_litt=3 & $48.0$ & $0.0$    & $42.0$ & $8.2$   & $6.0$  \\ \hline
    move=8, det\_area=3, k\_litt=3 & $48.0$ & $1.6$    & $39.5$ & $17.2$  & $8.5$  \\ \hline
    \end{tabular}
    \caption{Evaluation results.}
    \label{tab:eval}
\end{table}

The results (see Table~\ref{tab:eval}) clearly show that in all
considered situations the new ROS2-BDI framework equipped with
continual planning requires on average a smaller number of moves to
dispose of all the litter items.
The results also show that in the first six situations (where one
litter item is detected while moving) the improvements are higher.
This is due to the fact, that in the previous ROS2-BDI framework each
time encounters a failure in the execution for an obstacle or the
discovery of a new litter item needs to compute a full plan from its
current position, while the new one is equipped with continual planning
can easily adapt the not yet executed plans to avoid the obstacle and
to dispose of the newly detected litter items.
The results also show that when the recycling agent is able to observe
things within a proper anticipation window (i.e., not too late, not
too early that's \texttt{det\_area}=2 for the scenario), and there is
not much movement and we see the greatest improvement. This is due to the ability of the new ROS2-BDI approach to modify the solution plan to also fulfil the goal of disposing of the newly discovered litter item.
However, when the movement rate increases and the detection area is not
optimal, the improvements reduce. For instance, when
\texttt{det\_area}=1, the recycling agent discovers the new litter
item too late, going toward a sub-optimal solution and bumping more
frequently into occupied cells, due to its limited knowledge of the
current world status.
When \texttt{det\_area}=3, a "ping-pong" behaviour can potentially
happen; this is due to the fact that the agent has to pass through a large set of cells, and because of the broad detection area, the computed plan might be invalidated since the agent may discover one of
the locations to visit with that plan being occupied (despite the
location might be free when effectively reached). In this case, a new
plan to avoid the detected obstacle is generated. This problem could
be alleviated by introducing an additional functionality that instead
of simulating the whole plan, only the first $N$ actions are
simulated, and if no invalidation is detected, then continue the
execution thus delaying the replanning. (This functionality is left
for future development.)
The results show also that similar considerations hold when all the
three litter items are already known since the beginning (last six
rows in Table~\ref{tab:eval}). Here the new ROS2-BDI framework allows
for easily adapting the plan in execution to avoid newly discovered
obstacles.

%% file: sections/related_work.tex
The problem of designing complex and intelligent autonomous
architectures for the robotic setting has been the subject of several
works~\cite{DBLP:journals/aepia/GottifrediTCGS10,breemen2013,DBLP:conf/ecai/ScalaT14,duffy1999social,DBLP:conf/woa/AlzettaG19,cognitao,DBLP:journals/cogsr/BustosMBRGM19,DBLP:conf/aips/CashmoreFLMRCPH15,DBLP:conf/iros/0001CMR21}. All
these works suffer of severe limitation.
Some are not addressing autonomous deliberation
(e.g. \cite{DBLP:conf/woa/AlzettaG19}) and rely on pre-loaded
plans.
Others like e.g. CogniTAO~\cite{cognitao},
CORTEX~\cite{DBLP:journals/cogsr/BustosMBRGM19},
SkiROS2~\cite{DBLP:conf/taros/PolydorosGRNK16},
ROSPlan~\cite{DBLP:conf/aips/CashmoreFLMRCPH15}, and
PlanSys2~\cite{DBLP:conf/iros/0001CMR21} have the ability to
automatically generate new plans for handling contingencies. However,
all these frameworks first generate a full plan for the given goal
before starting its execution. When the execution of a plan fails
because of e.g. an unexpected contingency, a new full plan is
generated and then executed. Moreover, all these frameworks lack of
several BDI capabilities like e.g. detection-reaction, multi-agent
interaction mechanisms, and automatic generation of new desires.
The ROS2-BDI framework~\cite{PAAMS22} overcomes many limitations of
all the previous approaches by providing BDI capabilities.  However,
even in this framework, the execution of a plan happens only after the
generation of a complete plan to achieve a goal. In our framework, we
extend the ROS2-BDI framework to integrate continual planning while
executing the plan, thus enabling an agent to i) start acting before a complete plan has been generated; ii) quickly adapt the agent's
intentions to possible contingencies that may happen while executing.

In \cite{DBLP:journals/ai/PatraMGNT21} a framework for interleaving
action and planning with operational models has been proposed:
whenever a task during execution needs to be refined a planner is
invoked. Similarly to our case, the planner searches from a given
depth, it returns the computed task refinement and starts its
execution. In parallel, it continues the search to refine the given
task. Differently from our case, they consider instantaneous actions,
and the approach has not been used within a BDI framework.

In~\cite{concurr_pexec_arobot} it has been proposed a software system
integrating perception, planning, and real-time control where planning
and execution can occur in parallel: while executing an action, the
motion planning of the immediate next action is computed concurrently.
A similar approach was also presented in \cite{895924}.
A framework for reactive run-time composition (with no search to guide
the agent towards goal achievement) of pre-defined skills has been
proposed in \cite{9473021}.

Other researchers addressed the problem of interleaving planning and
acting with stochastic approaches, focusing more on theoretical
aspects, such as the completeness of the solution, and less on the
scalability and feasibility of such approaches to real-world use
cases~\cite{rt_concurr_pexec_stoc_domains,prob_planning_comp_hard,cond_plan_theorem_prover}.
In~\cite{DBLP:journals/ai/BertoliCRT06} it has been proposed an approach based on planning under partial observability that allows for
generating conditional plans that depending on the sensing execute
a different course of instantaneous actions. Such plans can then be
encoded as reconfigurable behaviour trees~\cite{reconfig_bt} to
facilitate their execution in the robotic setting.
However, all these works first compute a plan 
and then they try to execute it and do re-planning in case of
execution failure.

\cite{cont_planning_acting_ma}
presents an algorithm for continual planning where in the planning
domain one has to complement the classical specification of the
actions with constructs to describe why and when the agent should
switch between planning and acting. Their algorithm enables agents to
deliberately postpone parts of their planning process and actively
gather missing information relevant for subsequent refinement of the
plan. This work is limited to instantaneous actions,
assumes no concurrent execution of actions, requires a
specialized search algorithm and a modification of the PDDL problem to
include special constructs to specify the strategies governing the
switch between planning and acting. In our approach, we overcome their
limitations by considering durative actions, and relying on PDDL
2.1 that is used by a state-of-the-art planning algorithm
slightly modified to realize the continual planning and execution
through interaction with the whole BDI architecture.


%% file: sections/conclusion.tex

We proposed a novel architecture for continual temporal planning in
the BDI model. The architecture has been implemented as a significant extension of ROS2-BDI. We carried out an experimental validation
to compare the continual planning with the previous ROS2-BDI framework within a scenario simulated by interfacing the developed framework with the Webots simulation environment.
The experiments showed improved results compared to ROS2-BDI.

This work constitutes the basis for a variety of future work such as:
\begin{enumerate*}[label=\roman*)]
\item  Extension of this work to a full-fledged real-time framework to guarantee schedulability in terms of computational capacity on the line of~\cite{DBLP:conf/woa/AlzettaG19,ijcai2022};
\item Improving the forecasting plan failure through more sophisticated prediction mechanisms based on e.g., machine learning or statistical reasoning; 
\item Capability of storing for later efficient reuse plans synthesized in a previous run;
\item  Considering dynamic search interval that auto-adapt to the world dynamic;
\item Extending validation in a multi-agent scenario, considering cooperative versus competitive behaviours;
\item Extending the support to alternative planning approaches
  including non-deterministic~\cite{DBLP:journals/ai/BertoliCRT06,DBLP:journals/ai/CimattiDMRS18} and probabilistic planning.
\end{enumerate*}